\documentclass[11pt]{article}
\usepackage{acl}

\usepackage{times}
\usepackage{latexsym}

\usepackage[T1]{fontenc}

\usepackage[utf8]{inputenc}

\usepackage{microtype}

\usepackage{CJKutf8}

\usepackage{graphicx}
\usepackage{tabularx}
\usepackage{soul}

\usepackage{epstopdf}
\usepackage{xstring}

\usepackage{color}

\usepackage{enumitem}
\usepackage{qtree}
\usepackage{amsmath}
\usepackage{mathrsfs}
\usepackage{amssymb}
\usepackage{gb4e}
\noautomath
\usepackage{algorithm}
\usepackage{algorithmic}
\usepackage{qtree}
\usepackage{comment}

\title{CJaFr-v3 : A Freely Available Filtered Japanese-French Aligned Corpus}
\author{Raoul Blin and Fabien Cromières\\
        cnrs-crlao , independent\\
        blin@ehess.fr , fabien.cromieres@gmail.com}

\begin{document}
\maketitle

\begin{abstract}
We present a free Japanese-French parallel corpus. It includes 15M aligned segments and is obtained by compiling and filtering several existing resources.  In this paper, we describe the existing resources, their quantity and quality, the filtering we applied to improve the quality of the corpus, and the content of the ready-to-use corpus. We also evaluate the usefulness of this corpus and the quality of our filtering by training and evaluating some standard MT systems with it.
\end{abstract}


\section{Introduction}

Until recently, there was no large Japanese-French corpus (several million bisegments or more), \textit{freely available}. A fortiori, there was no ready-to-use for translation. 
The only corpus whose quality has been evaluated did not exceed 440K bisegments \citep{kitajafr-v1.1.0.0}, mainly constituted of TED corpus \citep{TED2020}. This placed Japanese-French among the low-resource language pairs. 
This was paradoxical because Japanese on the one hand and French on the other are languages that are very present in automatic processing. They are associated with other languages in very large corpora (for example, French-European languages, Japanese-English). There are also countless monolingual corpora of all sizes. Moreover, Japanese and French are languages with a large number of speakers (over 120M Japanese speakers; over 200M French speakers\footnote{http://observatoire.francophonie.org/ qui-parle-francais-dans-le-monde/; 2021-01-10}). The potential number of users of a bilingual corpus is therefore not negligible.  

Thanks to the availability of relatively new crawled corpora of several million bisegments, the situation has changed significantly and made it possible to build a large corpus. Unfortunately, the crawled corpora available are more or less noisy and cannot be used as is.




Our goal in this work is to build and provide a large (15M bisegments), freely available, Japanese-French corpus, CJaFr-v3 
\footnote{http://crlao.ehess.fr/rblin/tajafr.php\#cjafr}
. 
It is obtained by compiling and filtering
existing resources, including crawled corpora.
We also provide a standard split of the data to provide a ready-to-use corpus. In this way, we hope to significantly facilitate research on the translation and analysis of the Japanese-French pair.

This article is structured as follows. 
In section 2 and 3, we list the usable data.
In section 4, we describe the filters and processes we applied to the crawled corpora to reduce the noise.
In section 5, we evaluate the effect of the filtering by training two Transformer models on the crawled data. We observe that the filtered version provide a better model.
In section 6, we describe the structure of the ready-to-use dataset. 

\section{Available corpora}

Today, there are several large, freely available Japanese-French corpora. 
The OPUS project \citep{tiedemann_opus_2004} provides the largest number of corpora (about 20 corpora, $\approx$29M sentences). Some of the corpora are also available on the site of their original producers (e.g. WikiMatrix \citep{wikimatrix} at https://ai.facebook.com/blog/ wikimatrix/). 
The crawled corpora WikiMatrix, CCMatrix \citep{ccmatrix}, MultiCCAligned \citep{CCAligned} are the largest (21M bisegments in all). They are aligned using English as a pivot language. 
OPUS also provides TED and Tatoeba \footnote{tatoeba.org} corpora, all translated by humans. Tatoeba is not large and the sentences are relatively short. TED is not a native French or Japanese corpus. It is largely (if not entirely) translated from English into French and Japanese. 

Among the corpora available in OPUS, we excluded very small corpora containing less than 0.1M sentences (e.g. EUbookshop). 
Indeed, given the expected benefit, the time required for correction was disproportionate.
Six corpora (GNOME, KDE4, etc.) from computer documentation were excluded too. These are more often comparable corpora than aligned corpora. The vocabulary includes a large amount of English technical vocabulary. To avoid repetition of bisegments, we did not exploit any sub-corpora (e.g. XLEnt, extracted from WikiMatrix; CCAligned from MultiCCaligned etc). Finally, QED is rather a comparable corpus. Its filtering was too complex and it did not allow us to obtain a sufficient number of bisegments. 
\citep{blin_ALR2018} noted that the bisegments in Open-subtitles \citep{OpenSubtitles2016} are not correctly aligned. But above all, it is a corpus of very short dialogues which require special treatment because the two oral languages are very different.

Several small corpora of a few tens of thousands of bisegments have been added to those from OPUS. The Jibiki project\footnote{jibiki.fr}
provides the Cesselin, an old Japanese-French dictionary, published in 1940 and that has been OCRed \cite{cesselinmangeot}, and the  `Status of the International Criminal Court'
\footnote{Japanese version: www1.doshisha.ac.jp/~karai/intlaw/ docs/ icc.htm\\
French version: childrenandarmedconflict.un.org/key documents/french/ romestatuteofthe7.html
}. 

The Tajafr project \citep{kitajafr-v1.1.0.0} provides several small specialised corpora. Among them, two collections of titles (from French embassy website and press). 
Titles are interesting because of their very specific syntax in Japanese. Paradoxically, while they occupy a visible place in many types of texts, they are completely neglected in natural language processing and linguistics. 
Press headlines (natively in Japanese) are provided with two translations (litteral and non-litteral).
Tajafr also provides a collection of Japanese idiomatic expressions translated into French and PUD\footnote{Test corpus used at CoNL 2017 shared task on parsing Universal Dependencies. lindat.mff.cuni.cz/ repository/ xmlui/ handle/11234/1-2184}, used as test corpus.

\section{Quality}
The content and the form of the sub-corpora of CJaFr are very diverse. It contains oral presentations (TED), familiar expressions (tatoeba), legal texts (CCI, tax convention), encyclopedic style (Wikipedia) etc. We assume that the crawled corpora are by nature very diverse and lacking uniformity. 

We assume that the diversity is quantitatively perceptible through two measures at least: the variations of the segment lengths, and the richness of the vocabulary 
We measure the length of the segments and the standard deviation of the length/average-length ratio. To measure and compare the richness of the vocabulary, for each sub-corpus we randomly extract the same number $(N=1000)$ of words and measure the vocabulary ($V$) for this sample. The values are normalised ($V/N$). Data are provided in (Tab.\ref{tab_souscorpustaille}).
 
\begin{table*}
\centering
\begin{tabular}{lccrrrrcrr}
               & prod & cotext & \multicolumn{2}{|c}{segments} &\multicolumn{3}{|c|}{\#mots}   & \multicolumn{2}{c}{voc}      \\ \hline
Corpus         &          &           &    \#   &  L ($\sigma$)$^{(4)}$    &ja    & fr       & fr\%$^{(3)}$        &ja & fr  \\ \hline\hline
ccmatrix       & c & n/a    &         & 9.2 (0.71)       & 118.1M & 84.0M  &  60& .442 & .622 \\ 
multiccaligned & c & n/a    &         & 14.6 (1.09)      & 103.9M & 74.3M & 64&  .480 & .623 \\ 
wikimatrix     & c & n/a    &        &  16.11 (0.58)      & 27.8M & 19.8M & 64 & .434 & .604 \\ 
ted            & t & yes      &        &  20.83 (0.73)    & 8.5M & 6.7M & &  .418 & .523 \\ 
bible          & t &   yes    &       &  20.83 (0.45)    & 11.2M & 0.7M    &                 & .342 & .488 \\ 
tatoeba        & t &no &             & 8.93 (0.51)       & 426,903 & 317,545 & &  .367 & .531 \\ 
int.court law  & t &yes &             & 31.16 (0.91)    & 45,662 & 32,911 &  & .312 & .418 \\ 
tax law        & t &yes   &             & 45.81 (0.96)  & 21,475 & 16,215 &  & .250 & .365 \\ 
titles\_embassy & t &no &             & 10.67 (0.43)     & 34,437 & 18,091 &  & .358 & .470 \\
locutions      & t &no &             & 6.32 (1.11)       & 6,765 & 6,667      & & .383 & .514 \\ 
press\_head$^{(1)}$ & t &no &          & 16.30 (0.33)     & 7,994 & 6,476 & &  .418 & .571 \\ 
press\_head$^{(2)}$  & t &no &         & 17.73 (0.30)    & 7,994 & 5,778 &  & .418 & .578 \\
\hline
ext            & c & n/a    &  14M    & 12.64 (\textbf{0.91})    &233.8M & 165.5M &  & \textbf{.478} & \textbf{.608} \\ 
cesselin       & t & no &   100,420     & 4.18 (0.79)     & 341,891 & 446,957 & &  .433 & .479 \\ 
train          & t & n/a &  423,975   & 20.30 (0.74)    &  9.9M & 9.2M &  & .409 & .454 \\ 
val            & t & n/a &    2,964   & 24.00 (0.79)      & 66,030 & 58,013 & &  .416 & .426 \\ 
ted.test       & t & yes & 2,929        & 18.02 (0.70)    & 69,510 & 65,436 & &  .409 & .477 \\ 
PUD            & t & no &  1,000        & 24.00 (0.41)   & 29,460 & 25,269 &  & .435 & .517 \\ 
test           & t & n/a &   2,943    & 19.87 (0.72)     & 67,965 & 62,971 & & .439 & .469 \\ 
\hline
\end{tabular}
\caption{Characteristics of sub-corpora after filtering and before BPE segmentation; Richness of the vocabulary (sample of 1000 words); `prod'uction: (c)rawled corpus (t)ranslated corpus; `press headlines': (1) literal translation; (2) non literal translation; (3) Remaining text after filtering (\% of words in French); (4) standard deviation of length ratios} 
\label{tab_souscorpustaille}
\end{table*}

The majority of the available corpora have the same shortcomings. First, most of them do not contain native French or Japanese texts. They are often synthetic corpora, translated from English. This is typically the case with TED. 
Synthetic texts are not in themselves a problem if the translation is of good quality. On the other hand, it is very possible that many expressions typical of each of the two languages, Japanese and French, are missing and that more anglicisms are present.
Second, because of the alignment by segments, mostly sentences, it is not possible to know the cotext
\footnote{As a reminder, ``cotext'' (not ``context'') refers to the sentences that immediately precede.}
. This is a problem especially for retrieving elided elements in Japanese (a pro-drop language). In (Tab.\ref{tab_souscorpustaille}) we indicate which corpora contain sentences that originally do not have cotext. 


The crawled corpora represent the main part of the available corpora. It is necessary to indicate their specific defects. 
\begin{enumerate}
\item Presence of repeated bisegments.
\item Unusual strings of symbols (e.g. `;;;;;;').
\item Many automatically translated texts with poor quality (e.g. incorrect syntax, incomprehensible meaning, etc.
\footnote{Example: 
\begin{CJK}{UTF8}{min}
彼らは強力であり、非常に遅いために非常に抵抗性の動物で有名な、しかし。
\end{CJK}
\\
Ils sont forts et des animaux très résistants célèbre pour être très lent, mais.\\
\textit{``$\approx$ They are strong and some animals very resistant famous to be very slow, but.''}
})
.This phenomenon had already been raised by \citep{mosesmtmarketreport2015}.

\item No standardisation (e.g. encoding of  characters; format of numerals etc.). 
%
\item Neither the alignment nor the quality of translations is validated by humans. 
\end{enumerate}

\section{Bisegment filtering and processing}

We proceeded to a filtering, according to the following criteria. These criteria were decided by successive approximations, in order to obtain the best translation model. Some of the criteria are already used in filtering toolbox (see for example \citep{opusfilter-2020}).
\begin{enumerate}
\item Elimination of repeated bisegments. 
\item Filtering by structure matches (parentheses): the number of opening and closing symbols must be identical in both segments. 
\item Limitation of the length of the segments(350 bytes). 
\item The size ratio between the two segments in a bisegment must not exceed 1 to 3. 
\item Elimination of bisegments that contain characters strings we consider as not belonging to canonical expressions (e.g. `` ;;;;; ''); more than two characters among ``$\backslash\slash$:!?\$'', or any logical characters, iconic graphems (flags, emoticons etc.); more than 20 uppercase characters, and the same for digits.
\end{enumerate}
We note that only $\approx$60\% of the crawled corpus remains after filtering.

The Cesselin dictionary received a specific processing. We used only the translated examples, not the lexical entries. We have systematically eliminated examples containing incongruous characters generated by the OCR processing. The Japanese language is typical of the 19th and beginning of 20th century. It differs from the contemporary language. This is particularly true of verbal and adjectival conjugations. We have roughly modernised them. The conjugations have been replaced by the contemporary forms (ka-u-te (``buy'') $\rightarrow$ ka-tte, etc.) old kana are modernised (e.g. \begin{CJK}{UTF8}{min}ゐ\end{CJK} $\rightarrow$ \begin{CJK}{UTF8}{min}い\end{CJK}). However, some inappropriate forms remain and the resulting text cannot be considered completely modernised. 
Because of this, after processing, this corpus is usable for translating from Japanese into French, but is not reliable for French into Japanese.

\section{Evaluation of filtering}

We have evaluated the effect of the  filtering. We trained a model on a sample of the raw crawled corpus 
and a model on a sample of the filtered crawled corpus. The sample size is set to $\approx20M$ Japanese words. Japanese is segmented using mecab \citep{kudo_mecab_2006} and Unidic-cwj \citep{unidic_cwj2017_en} dictionary. 
French is tokenised and truecased with the standard Moses tools. 
Japanese and French are segmented using SentencePiece \citep{sentencepiece} (BPE segmentation) with a vocabulary set to 8000 for both. The BPE segmentation model is trained and applied independently on each corpus sample (we then have four segmentation models). 

The training is executed with Opennmt-py.2.0.0 \citep{OpenNMT}
\footnote{The hyperparameters are those suggested in \texttt{opennmt.net/OpenNMT-py/FAQ.html\#how-do-i
-use-the-transformer-model; 2021/06/01}; batch size= 2048; word vec size = 256}. 
We perform two trainings for each corpus and calculate the average. 

The translation was evaluated after removing segmentation and separating the punctuation. 
The evaluation is carried out on 2 test corpus: PUD and ted.test. 
The scores used are BLEU\footnote{Calculated using multi-BLEU \texttt{www.statmt.org/wmt06/shared-task/ multi-bleu.perl}, default settings} and c6+w2-avgF2 \citep{chrf}. 

The results (Tab.\ref{tabcomp}) show that training over the filtered corpus leads to a globally better translation quality than using the raw corpus. 
However, the results are not uniform. The values are significantly better for ja>fr, but slightly negative for fr>ja. 

\begin{table}
\begin{centering}
\begin{tabular}{lccrrrr}
   &        &     \multicolumn{2}{c}{raw}   &  \multicolumn{2}{c}{filtered$^b$}   \\ 
lg & corpus &     BLEU & chrF   & BLEU     & chrF  \\ \hline
j>f & PUD   & 8.06 & 35.80 & 1.96 & 4.21 \\ 
j>f & ted   & 5.21 & 30.42 & 1.48 & 2.13 \\ 
j>f  & ${(a)}$   & 6.63 & 33.11 & 1.72 & 3.17 \\ 
f>j & PUD   & 13.73 & 28.48 & -0.84 & -1.31 \\ 
f>j & ted   & 10.33 & 20.77 & 0.22 & -0.16 \\ 
f>j &  ${(a)}$   & 12.03 & 24.63 & -0.31 & -0.73 \\ 
${(a)}$  & PUD    & 10.90 & 32.14 & 0.56 & 1.45 \\ 
${(a)}$  & ted   & 7.77 & 25.60 & 0.85 & 0.98 \\ 
\multicolumn{2}{l}{Overall}       & 9.33 & 28.87 & +0.70 & +1.22 \\ \hline
\end{tabular}
\caption{Test scores ( (a) both language pairs or corpora (b) difference: filtered score - raw score)
\label{tabcomp}}
\end{centering}
\end{table}

\section{Construction of CJaFr}

CJaFr (version 3) is a package of corpora. The first set of data is constituted of the original corpora (including filtered crawled corpora). It already has been described in sections above. The second set of data is the ready-to-use corpus. 
It is structured in two parts (see also table \ref{tab_souscorpustaille}): a core and an extension. 
The core was built by gathering corpora produced by translation, not webcrawling. 
The quantitatively most important part consists of TED. It also contains the legal corpora, locutions, titles, Tatoeba etc. This core is classically decomposed into three sub-corpora: a fine-tuning corpus (3000 bisegments) and a test corpus (3000 bisegments). It contains two more test corpora: a sample of 3000 TED bisegments and PUD. 
The extension is composed of the filtered crawled corpora and the Cesselin.

\section{Conclusion}

To our knowledge, CJaFr corpus (v3) is the only freely available ready-to-use Japanese-French corpus of its size.
The core is made of corpora translated by human. The extension part is obtained by filtering crawled corpora with simple techniques. 
The size of the core remains much smaller than the size of the translated (vs crawled) corpora available for English-Japanese in particular. It is also smaller than the theoretical size of  human learning corpus (see a discussion in \citep{corpusfrugal}). 
However, adding all or part of the extension allows us to work comfortably with neural networks, and we hope it will facilitate new works for this language pair.


The crawled corpora are an undeniably welcome resource. However, at the end of this compilation work, we wonder about the interest of crawled corpora today, for a pair of languages like Japanese and French. On one hand, the quality of the crawled corpora is very variable. The resources are difficult to control (quality, content, copyright) and contain a large amount of poor quality machine translations, made by unknown machine translators. On the other hand, it exists reasonably good (commercial) translators.
We suspect that it would be preferable to build synthetic corpora with such translators: take a native target corpus and generate a synthetic source corpus. Unlike the crawled corpus where both source and target texts are not checked, the interest is that at least, the target is attested to be of good quality.

%

\bibliography{anthology,acl_latex,bibcomplete,blin}
\bibliographystyle{acl_natbib}

\end{document}